\documentclass[conference]{IEEEtran}
\IEEEoverridecommandlockouts
\usepackage{tikz}
\usepackage{algorithmicx}
\usepackage{algorithm}
\usepackage{algpseudocode}
\usepackage{amsmath}
\usepackage{bm}
\usepackage{vector} 
\usepackage{amsfonts}
\usepackage{siunitx}
\usepackage{commath}
\usepackage[backend=biber,doi=false,style=ieee,natbib=true]{biblatex} 
\usepackage{lipsum}
\usepackage{nohyperref} 
\usepackage{cleveref}
\usepackage{multirow}

\usepackage{longtable,tabularx,booktabs}
\usepackage[flushleft]{threeparttable}
\usepackage[keeplastbox]{flushend}

\DeclareMathOperator{\Normal}{\mathcal{N}}

\DeclareMathOperator{\sign}{sign}

\DeclareMathOperator*{\argmax}{arg\,max}

\newcommand{\mat}[1]{\vect{#1}}
\renewcommand{\vec}[1]{\vect{#1}}
\newcommand{\floor}[1]{\lfloor #1 \rfloor}

\definecolor{pastelMagenta}{HTML}{FF48CF}
\definecolor{pastelPurple}{HTML}{8770FE}
\definecolor{pastelBlue}{HTML}{1BA1EA}
\definecolor{pastelSeaGreen}{HTML}{14B57F}
\definecolor{pastelGreen}{HTML}{3EAA0D}
\definecolor{pastelOrange}{HTML}{C38D09}
\definecolor{pastelRed}{HTML}{F5615C}

\algtext*{EndLoop} 
\algtext*{EndIf}
\algtext*{EndFunction}

\usetikzlibrary{calc}
\usetikzlibrary{shapes.geometric}
\usetikzlibrary{external}
\usetikzlibrary{patterns}
\usetikzlibrary{shapes,arrows,fit}
\usetikzlibrary{positioning}
\usetikzlibrary{arrows.meta, calc, shapes}
\usetikzlibrary{graphs}
\usetikzlibrary{decorations.pathmorphing}
\usetikzlibrary{decorations.pathreplacing}

\newcommand{\citefull}[1]{\citeauthor{#1}~\cite{#1}}
\newcommand{\smallcaps}[1]{\textsc{#1}} 

\begin{document}

\title{Adaptive Stress Testing of Trajectory Predictions in Flight Management Systems}


\author{\IEEEauthorblockN{Robert J. Moss\IEEEauthorrefmark{1}, Ritchie Lee\IEEEauthorrefmark{2}, Nicholas Visser\IEEEauthorrefmark{3}, Joachim Hochwarth\IEEEauthorrefmark{3}, James G. Lopez\IEEEauthorrefmark{4}, and Mykel J. Kochenderfer\IEEEauthorrefmark{1}}
\IEEEauthorblockA{\IEEEauthorrefmark{1}Stanford Intelligent Systems Laboratory, Stanford University, Stanford, CA, 94305\\
\IEEEauthorrefmark{2}NASA Ames Research Center, Moffett Field, CA, 94035\\
\IEEEauthorrefmark{3}GE Aviation, Grand Rapids, MI, 49512\\
\IEEEauthorrefmark{4}GE Global Research, Niskayuna, NY, 12309
}}

\maketitle

\begin{abstract}  
To find failure events and their likelihoods in flight-critical systems, we investigate the use of an advanced black-box stress testing approach called adaptive stress testing.
We analyze a trajectory predictor from a developmental commercial flight management system which takes as input a collection of lateral waypoints and en-route environmental conditions.
Our aim is to search for failure events relating to inconsistencies in the predicted lateral trajectories.
The intention of this work is to find likely failures and report them back to the developers so they can address and potentially resolve shortcomings of the system before deployment. 
To improve search performance, this work extends the adaptive stress testing formulation to be applied more generally to sequential decision-making problems with episodic reward by collecting the state transitions during the search and evaluating at the end of the simulated rollout. 
We use a modified Monte Carlo tree search algorithm with progressive widening as our adversarial reinforcement learner.
The performance is compared to direct Monte Carlo simulations and to the cross-entropy method as an alternative importance sampling baseline.
The goal is to find potential problems otherwise not found by traditional requirements-based testing.
Results indicate that our adaptive stress testing approach finds more failures and finds failures with higher likelihood relative to the baseline approaches.
\end{abstract}


\section{Introduction}
\label{sec:introduction}

A primary function of aircraft flight management systems (FMS) is to provide guidance in the form of navigational waypoints between origin and destination airports.
A trajectory predictor is the subsystem that provides trajectories to the guidance subsystem that commands the autopilot.
Failures within the trajectory prediction system can occur if the output waypoints are unreachable given the physical limitations of the aircraft, or if there are problems with the implementation or design of the software. 
The goal of this work is to find likely failure cases before system deployment so that the engineers can resolve or address potential problems in the system. 

Traditionally, large-scale Monte Carlo testing is used to generate these failure cases \cite{monte_carlo}. However, Monte Carlo testing can be inadequate for large input spaces with rare failure events \cite{mc_limitations}.
We investigate the use of adaptive stress testing (AST), an advanced black-box stress testing approach that has been successfully applied to find failures in safety-critical systems \cite{ast_acasx,ast_av,lee2018differential,ast_safety}. Adaptive stress testing is a method that uses reinforcement learning to adversarially search for rare failure events in sequential decision-making systems \cite{lee_thesis}. This work applies the AST approach to trajectory prediction systems to efficiently find failure events and their likelihoods.

The trajectory prediction system is treated as a black-box simulator and AST controls the selection of waypoints and other environmental input parameters.
Monte Carlo tree search (MCTS) with progressive widening is used to explore the possible trajectories and a notion of ``miss distance'' to a failure event is used to help guide the search.
Transition probabilities between states are also used to guide the search towards the most likely failures.
In traditional AST formulations, a sequential decision-making problem is assumed, but our open-looped trajectory predictor does not provide a sequence of decisions, but rather generates trajectories based on complete flight plans.
We regard each waypoint in the flight plan as a step in the sequence.
Although we can forcibly fit the problem to a strict sequential framework by calling the system with partial flight plans, this is prohibitively expensive and unnecessary.
Instead, we collect the intermediate states and actions and only evaluate the system at the end of a simulated rollout, then back-propagate the reward to unevaluated parts of the tree.
We can do this due to the structure of the AST reward function.
Therefore, this work extends the AST approach to be applied more generally to sequential decision-making problems with episodic reward (i.e. rewards accumulated at the end of an episode with intermediate rewards of zero).

We analyze a trajectory predictor from a developmental commercial FMS which takes as input winds aloft, origin and destination airports, and a collection the lateral waypoints.
The trajectory predictor outputs the discrete-time controls that determine translational motion which would be input to the FMS.
Within the FMS, the trajectories are passed to the guidance subsystem which determines how to command the autopilot.
Although we focus on arc length discrepancies as the primary failure mode (described in \Cref{sec:application}), this work can be easily extended to other failure events.
The performance of AST is compared to two baselines: direct Monte Carlo simulation and the cross-entropy method.
Current failure assessment is performed exhaustively over a navigational database of predefined aircraft routes defined by lateral waypoints.
This testing method is used during development while requirement-based testing is used for final system certification following RTCA DO-178C \cite{do178c}.
The intention of testing during development is to find failures otherwise not covered by requirements-based tests.
As a comparison of developmental testing approaches, we sample routes from the navigational database as another baseline to relate the simulation-based approach to the standard navigational database approach.
Experiments were run to generate likely failure events that are provided to the developers of the trajectory predictor to analyze and resolve potential shortcomings of the system.

AST has been successfully applied to safety-critical systems such as aircraft collision avoidance systems  \cite{ast_acasx,lee2018differential} and autonomous vehicles  \cite{ast_av}. 
\citefull{ast_traj_plan} applied AST to trajectory planning systems, but as will be discussed further in \Cref{sec:approach_ast}, that problem had access to the full FMS, thus fitting the model of traditional AST.
They looked at runtime behavior of the FMS on a simulated aircraft, where the disturbances were added as sensor noise.
In our work, we do not rely on simulating aircraft dynamics and use input waypoints as the disturbances.

Other work has been proposed to efficiently search for failures in black-box cyber-physical systems \cite{corso2020survey}. \citefull{trustworthy_ai} propose an importance sampling approach to find rare failure events to assess autonomous vehicle safety.
Their work is similar to AST but relies on an accurate importance distribution.
Other approaches formulate the problem of falsification as an optimization problem and solve it using Bayesian optimization \cite{bayes_opt}, simulated annealing \cite{abbas2013probabilistic,aerts2018temporal}, or rapidly-exploring random trees  \cite{rrts}.
Each of these approaches are formulated as a classical optimization problem and use different techniques to search the input space for failures---although there is no likelihood estimation for a given falsifying input.
A different approach altogether uses an existing set of falsifying inputs to bootstrap the search for neighboring failures \cite{bootstrap}.
That work relies on existing counterexamples to base the neighboring search upon.
Our work addresses falsification of sequential systems with episodic reward and includes most likely failure analysis.

This paper is organized as follows. 
Section \ref{sec:background} provides necessary background of AST and commercial aircraft FMS. Section \ref{sec:approach} describes how this work extends existing AST approaches and modifications made to the MCTS algorithm. Section \ref{sec:implementation} details the implementation of this work using the Julia programming language and provides a description of the interface for AST to interact with black-box systems. Section \ref{sec:application} describes the simulation environment constructed for the FMS application and the failure events we are searching for. Section \ref{sec:experiments} describes the experiments and discusses the analysis of the results. Lastly, \Cref{sec:conclusion} provides a discussion on the conclusions from this work.

\section{Background} \label{sec:background}
This section describes background of the adaptive stress testing approach and flight management systems.

\subsection{Adaptive Stress Testing} \label{sec:background_ast}
Adaptive stress testing (AST) is a black-box approach to find rare failure events in cyber-physical systems \cite{ast_acasx,lee2018adaptive}.
The AST problem is formulated as a Markov decision process (MDP) and can be solved using reinforcement learning algorithms to guide the search towards likely failure events.
AST can also be formulated more generally as other sequential decision-making processes, such as a partially observable Markov decision process (POMDP) \cite{ast_acasx,ast_av,ast_traj_plan}.

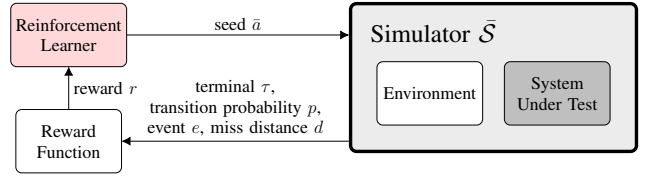
\begin{figure}[!t]
\centering
\resizebox{0.95\columnwidth}{!}{\begin{tikzpicture}
  [
    node distance=1.5cm,
    every node/.style={font=\large},
    align=center
  ]
      \tikzset{
        >={Latex[width=2mm,length=2mm]},
        base/.style = {rectangle, rounded corners, draw=black,
                       minimum width=1cm, minimum height=1cm,
                       text centered},
        block/.style = {base, minimum width=2.5cm, minimum height=1.5cm},
        envstyle/.style = {block, fill=red!15},
        sutstyle/.style = {block, fill=gray!50}, 
        simstyle/.style = {base, thick, fill=gray!15, minimum width=4cm},
        metasimstyle/.style = {base, line width=0.07cm, fill=orange!80!gray!40, minimum width=4cm},
        learnerstyle/.style = {block, fill=red!15}, 
        rewardstyle/.style = {block, fill=white}, 
        componentstyle/.style = {block, fill=white}, 
        aircraftstyle/.style = {block, draw=none, fill=brown!50!red!30},
        casstyle/.style = {block, fill=orange!50},
    }
    \pgfdeclarelayer{metasimlayer}
    \pgfdeclarelayer{simlayer}
    \pgfdeclarelayer{inputlayer}
    \pgfdeclarelayer{blackboxlayer}
    \pgfsetlayers{metasimlayer,simlayer,blackboxlayer,inputlayer,main}

    \node (wptdist) [componentstyle] {Environment};
    \node (sut) [sutstyle, right=0.5cm of wptdist] {System\\Under Test};

    \begin{pgfonlayer}{simlayer}
        \path let \p1=(wptdist.north west), \p2=(sut.south east) in node (sim) [fit={($(\x1,\y1)+(-0.5cm,1.25cm)$) ($(\x2,\y2)+(0.5cm,-0.5cm)$)},simstyle,line width=0.08cm]{};
        \node at ($(sim.north west)+(2cm,-7mm)$) [font={\LARGE}] {Simulator $\mathcal{\bar{S}}$};
    \end{pgfonlayer}

    \node (learner) [learnerstyle, above left=0.25cm and 5.25cm of sim.west,xshift=0cm] {Reinforcement\\Learner};
    \node (reward) [rewardstyle, below=1cm of learner,xshift=0cm] {Reward\\Function};

    \draw[->] (learner) -- (sim.west |- learner) node[pos=0.5,above]{seed $\bar{a}$};
    \draw[->] (reward.north) -- (learner.south) node[pos=0.5,right]{reward $r$}; 
    \draw[<-] (reward) -- (sim.west |- reward) node[pos=0.5,above]{terminal $\tau$,\\transition probability $p$,\\event $e$, miss distance $d$};
\end{tikzpicture}}
\caption{Adaptive stress testing formulation.}
\label{fig:ast_mdp}
\end{figure}

\Cref{fig:ast_mdp} illustrates how the AST concept is formulated.
The system under test is treated as a black box while the simulator $\mathcal{\bar{S}}$ is treated as a gray box that passes the transition probability $p$, event indicator $e$, miss distance $d$, and termination state indicator $\tau$ to the reward function.
The AST problem is solved using an adversarial reinforcement learner that selects a random number generator seed $\bar{a}$ to indirectly control the SUT through the simulator.
In other types of simulators, AST could directly control input disturbances rather than seeds.
In prior work, Monte Carlo tree search and deep reinforcement learning have been used to solve the MDP \cite{ast_acasx, ast_av}.
The output of the AST process is a set of state trajectories deterministically controlled by a set of seeds. 
These seeds are used to deterministically playback the simulation starting from an initial state.

A necessary clarification is differentiating what is black-box with respect to the AST problem.
Given that the simulator must provide $\langle p, e, d, \tau \rangle$ to the AST reward function, the output needs access to the transition probabilities $p$ and terminal indication $\tau$ from the environment, but can determine the event indication $e$ and miss distance $d$ from the output of the SUT.
Thus, depending on the problem, the environment may be required to be white-box but the SUT is strictly black-box.

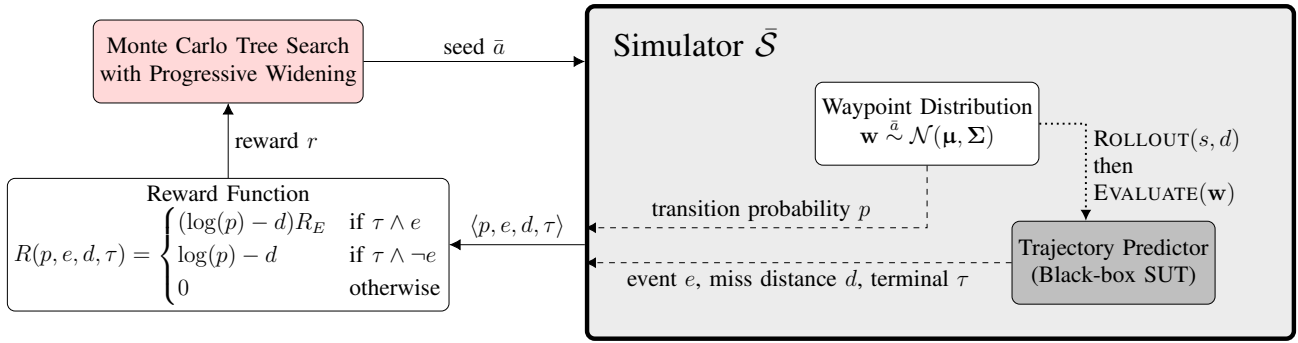
\begin{figure*}[!t]
  \centering
  \resizebox{0.95\textwidth}{!}{\begin{tikzpicture}
  [
    node distance=1.5cm,
    every node/.style={font=\large},
    align=center
  ]
      \tikzset{
        >={Latex[width=2mm,length=2mm]},
        base/.style = {rectangle, rounded corners, draw=black,
                       minimum width=1cm, minimum height=1cm,
                       text centered},
        block/.style = {base, minimum width=2.5cm, minimum height=1.5cm},
        envstyle/.style = {block, fill=red!15},
        sutstyle/.style = {block, fill=gray!50}, 
        simstyle/.style = {base, thick, fill=gray!15, minimum width=4cm},
        metasimstyle/.style = {base, line width=0.07cm, fill=orange!80!gray!40, minimum width=4cm},
        learnerstyle/.style = {block, fill=red!15}, 
        rewardstyle/.style = {block, fill=white}, 
        componentstyle/.style = {block, fill=white}, 
        aircraftstyle/.style = {block, draw=none, fill=brown!50!red!30},
        casstyle/.style = {block, fill=orange!50},
    }
    \pgfdeclarelayer{metasimlayer}
    \pgfdeclarelayer{simlayer}
    \pgfdeclarelayer{inputlayer}
    \pgfdeclarelayer{blackboxlayer}
    \pgfsetlayers{metasimlayer,simlayer,blackboxlayer,inputlayer,main}

    \node (wptdist) [componentstyle] {Waypoint Distribution\\$\vec{w} \overset{\bar{a}}{\sim} \mathcal{N}(\vec{\mu},\mat{\Sigma})$};
    \node (sut) [sutstyle, below right=1cm and -0.5cm of wptdist] {Trajectory Predictor\\ (Black-box SUT)};

    \draw[->,dotted,line width=1pt] (wptdist) -| ($(sut.north)+(-5mm,0mm)$) node[pos=0.725,right,align=left]{{\sc Rollout$(s,d)$}\\then\\{\sc Evaluate$(\vec{w})$}};

    \begin{pgfonlayer}{simlayer}
        \path let \p1=(wptdist.north west), \p2=(sut.south east) in node (sim) [fit={($(\x1,\y1)+(-4cm,1.25cm)$) ($(\x2,\y2)+(1.3cm,-0.5cm)$)},simstyle,line width=0.08cm]{};
        \node at ($(sim.north west)+(2cm,-7mm)$) [font={\LARGE}] {Simulator $\mathcal{\bar{S}}$};
    \end{pgfonlayer}

    \draw[->,dashed,line width=0.5pt] (wptdist.south) |- ($(sim.west)+(0mm,-1cm)$) node[pos=0.5,xshift=-3cm,above]{transition probability $p$};
    \draw[->,dashed,line width=0.5pt] (sut) -- (sut -| sim.west) node[pos=0.5,below]{event $e$, miss distance $d$, terminal $\tau$};

    \node (learner) [learnerstyle, above left=1.25cm and 4cm of sim.west,xshift=0cm] {Monte Carlo Tree Search\\with Progressive Widening};
    \node (reward) [rewardstyle, below=1.35cm of learner,xshift=0cm] {Reward Function\\[1mm]$R(p,e,d,\tau) = \begin{cases} 
    (\log(p) - d)R_E & \text{if } \tau \wedge e \\
    \log(p) - d & \text{if } \tau \wedge \neg e \\
    0 & \text{otherwise}
    \end{cases}$};

    \draw[->] (learner) -- (sim.west |- learner) node[pos=0.5,above]{seed $\bar{a}$};
    \draw[->] (reward.north) -- (learner.south) node[pos=0.5,right]{reward $r$};
    \draw[<-] (reward) -- (reward -| sim.west) node[pos=0.5,above]{$\langle p,e,d,\tau \rangle$};
\end{tikzpicture}}
  \caption{
    \label{fig:ast_mdp_modified}
    Modified adaptive stress testing formulation for the trajectory predictor with episodic reward. The simulation environment samples waypoints from a distribution and passes those waypoints as input to the SUT at the end of the rollout.
    The modified reward function is guided by both the severity and likelihood of the failure event. Information on the dashed lines is only provided to the reward function when the SUT is evaluated.
  }
\end{figure*}

The standard AST reward function is designed to guide the search towards failure events and to maximize the likelihoods of those events. It is also affected by the notion of miss distance $d$: a measure of ``closeness'' to a particular event. The miss distance helps guide the search towards failures to search efficiently. The standard reward function is given by:
\begin{equation}\label{eq:standard_reward}
  R(p,e,d,\tau)=\begin{cases} 
      R_E & \text{if } \tau \wedge e \\
      -d & \text{if } \tau \wedge \neg e\\
      \log(p) & \text{otherwise}
  \end{cases}
\end{equation}
The non-negative constant reward for finding an event is given by $R_E$ and is generally set to $0$.
The boolean $e$ indicates when an event has been found and the boolean $\tau$ indicates that the simulation is in a terminal state.
If the simulation terminates without finding an event, then the negative miss distance $-d$ is awarded to guide the search towards failure.
Otherwise, the log-likelihood of each state transition is used, denoted by $\log(p)$.
This term is used to maximize the likelihood of the overall trajectory and is designed to guide the search towards likely failures.
Recall that the goal of reinforcement learning is to maximize the expected sum of rewards \cite{sutton2018reinforcement}.
Using log-likelihood means we can maximize the summations, which is equivalent to maximizing the product of the likelihoods.

\subsection{Flight Management Systems}
Aircraft flight management systems (FMS) have been a critical part in reducing workload of pilots in commercial aircraft by contributing to in-flight automation  \cite{fms_workload}.
Major components of FMS include flight planning, navigation, guidance, performance optimization, and trajectory prediction \cite{fms}. 
This work focuses on the subsystem of the FMS that generates the aircraft trajectory given a flight plan.
This subsystem, called the trajectory predictor, provides deterministic trajectories based on an operator-defined input flight plan and estimates of environmental conditions.
Inputs include winds aloft, origin and destination airports, aircraft weight, cost index (a balance between cost of fuel and time of arrival), and a set of navigational waypoints that define the lateral route.
For our purposes, we focus on the winds aloft, origin and destination airports, and the placement of the lateral waypoints.
The trajectory predictor outputs the discrete-time controls that determine translational motion (i.e. both vertical and horizontal motion) which become input to the guidance subsystem of the FMS.
The trajectories may be processed before being passed to the guidance in the cases where a change in the lateral or vertical trajectory needs to be anticipated and controlled towards.
Therefore, the trajectory predictor is not a strictly sequential decision-making problem because the full sequence of lateral paths are deterministically constructed based solely on the inputs.

\section{Approach}
\label{sec:approach}

Several modifications were made to adapt AST for sequential decision-making systems with episodic reward.
Modifications to the standard reward function were made to guide the search towards severe failures as well as likely failures.
Modifications to the Monte Carlo tree search algorithm are also described; adapting the search algorithm for efficient SUT evaluations and to use progressive widening with a single deterministic next state (given the nature of controlling seeds).

\subsection{Adaptive Stress Testing for Episodic Reward Problems} \label{sec:approach_ast}

Traditionally, AST is used to steer sequential decision-making systems towards likely failures by controlling the seed used to sample environmental variables within the simulator at each time step (as seen in \Cref{fig:ast_mdp}).
The issue with applying this formulation directly to the trajectory predictor is that this system does not rely on sequential feedback to generate the trajectory and solely relies on its set of inputs. 
Therefore, we propose a modification to the AST formulation to abstract the sequential nature of the problem to the simulation environment.
In other words, we collect the state transitions during the search and evaluate the system at the end of the rollout.
This distinction can be seen in \Cref{fig:ast_mdp_modified}.
The transition probability $p$ is output from the environment and the event indication $e$, miss distance $d$, and terminal state indication $\tau$ are output from the black-box SUT, i.e. the trajectory predictor.
The 4-tuple $\langle p, e, d, \tau \rangle$ is passed as input to the reward function and the transition probability $p$ and miss distance $d$ are used to guide the search.

\begin{figure*}[!t]
\centering
\resizebox{\textwidth}{!}{
\tikzset{
    nodes={draw, circle}, >=latex, -, level distance=0.5in,
    every node/.style={draw=black, thin, minimum size=6mm},
    norm/.style={edge from parent/.style={black,thin,draw}},
    emph1/.style={edge from parent/.style={line width=0.8pt,draw}},
    emph2/.style={edge from parent/.style={line width=1.2pt,draw}},
    emph3/.style={edge from parent/.style={line width=1.4pt,draw}},
    emph4/.style={edge from parent/.style={line width=1.6pt,draw}},
    semiselected/.style={line width=0.8pt},
    selected/.style={line width=1pt},
    root/.style={label=\textsc{#1}},
    baseline=(selection-root.base),
    state/.style={circle, norm,-},
    action/.style={rectangle, norm},
    rollout-end/.style={rectangle, draw=none, minimum height=0mm},
    rollout-edge/.style={->, decorate, decoration={snake, amplitude=1mm, post length=10pt, segment length=12pt}, dotted, line cap=round, dash pattern=on 0pt off 2\pgflinewidth},
}
\begin{tikzpicture}
    \node [root=Selection, state, selected] (selection-root) {}
        child [emph1, ->] { node [action, selected] {}
            child [state] { node {} }
            child [state, emph2, ->] { node [selected] {} }
        }
        child [state] {node [action] {}}
        child [state] {node [action] {}
            child [state] { node {} }
            child [state] { node {} }
        };
\end{tikzpicture}
\hspace{1cm}
\begin{tikzpicture}
    \node [root=Expansion, state] {}
        child { node [action] {}
            child [state] { node {} }
            child [state] { node [semiselected] {}
                child [state, emph3] { node [action, selected] {} }
            }
        }
        child [state] {node [action] {}}
        child [state] {node [action] {}
            child [state] { node {} }
            child [state] { node {} }
        };
\end{tikzpicture}
\hspace{1cm}
\begin{tikzpicture}
    \node [root=Rollout, state] {}
        child { node [action] {}
            child [state] { node {} }
            child [state] { node {}
                child [state] { node [action, selected] {}
                    child [state, emph4, level distance=0.74in] { node [rollout-end] {$Q$}
                        edge from parent [rollout-edge]
                    }
                }
            }
        }
        child [state] {node [action] {}}
        child [state] {node [action] {}
            child [state] { node {} }
            child [state] { node {} }
        };
\end{tikzpicture}
\hspace{1cm}
\begin{tikzpicture}
    \node [root=Backpropagation, state, selected] {}
        child [<-,emph1] { node [action, selected] {}
            child [state] { node {} }
            child [state, emph2, <-] { node [selected] {}
                child [state, emph3, <-] { node [action, selected] {$Q$} }
            }
        }
        child [state] {node [action] {}}
        child [state] {node [action] {}
            child [state] { node {} }
            child [state] { node {} }
        };
\end{tikzpicture}}
\caption{The four steps of the Monte Carlo tree search algorithm.}
\label{fig:mcts}
\end{figure*}
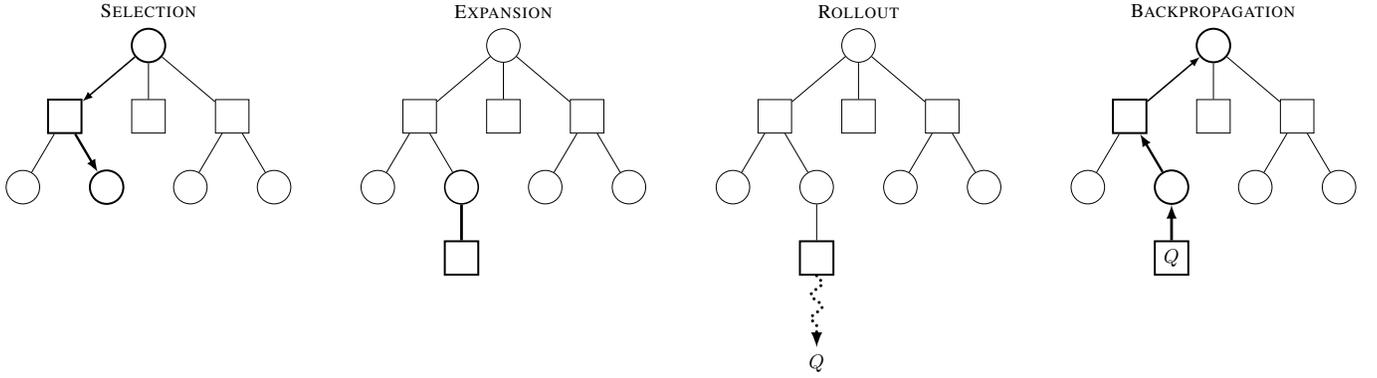

The standard AST reward function described in \Cref{eq:standard_reward} was modified to collect all rewards at the termination state and to incorporate a severity measurement when a failure event occurs.
Both the log-likelihood and miss distance $d$ are used throughout the search.
A multiplicative bonus $R_E$ is applied when a failure event occurs and we set $R_E=100$ for our experiments.
The modified reward function becomes:
\begin{equation}\label{eq:modified_reward}
  R(p,e,d,\tau) = \begin{cases} 
    (\log(p) - d)R_E & \text{if } \tau \wedge e \\
    \log(p) - d & \text{if } \tau \wedge \neg e \\
    0 & \text{otherwise}
  \end{cases}
\end{equation}
The transition probability $p$ is given by the probability of sampling a set of waypoints from a multivariate Gaussian distribution $\vec{w} \overset{\bar{a}}{\sim} \Normal(\vec{\mu},\mat{\Sigma})$ composed of waypoint direction and distance (relative to the previous waypoint) and wind direction and magnitude with mean vector $\vec\mu$ and covariance $\mat\Sigma$, deterministically controlled by the seed $\bar{a}$.
If an event is found, then the reward is the negative miss distance combined with the log-likelihood and adjusted by the multiplicative bonus $R_E$.
This modification is used to incorporate severity of an event gauged by the miss distance when an event is found.
If the simulation terminates without finding an event, no multiplicative bonus is applied.  
This reward function may not be suitable for certain AST problem formulations, but in \Cref{sec:application} we discuss how these modifications are applicable for the failure event and miss distance we investigate.

\begin{figure}[!b]
\centering
\resizebox{0.6\columnwidth}{!}{\begin{tikzpicture}[
  nodes={draw, circle}, -, level distance=0.35in, 
  every node/.style={circle, draw=black,thin, minimum size = 0.5cm},
  emph1/.style={edge from parent/.style={line width=0.8pt,draw}},
  emph2/.style={edge from parent/.style={line width=1.2pt,draw}},
  emph3/.style={edge from parent/.style={line width=1.4pt,draw}},
  emph4/.style={edge from parent/.style={line width=1.6pt,draw}},
  norm/.style={edge from parent/.style={black,thin,draw}},
  evaluate/.style={rectangle, draw=none, minimum height=1mm, edge from parent/.style={black,thin,draw,dotted}},
  action/.style={rectangle, minimum height=6mm, minimum width=6mm},
  evaluate-edge/.style={>=latex, ->, decorate, decoration={snake, amplitude=0.7mm, post length=10pt, segment length=12pt}, dotted, line cap=round, dash pattern=on 0pt off 2\pgflinewidth},
  level 2/.style={sibling distance=10mm},
]
\node{$s_0$}
  child [emph1] { node [norm, action] {$\bar{a}_1$}
    child [emph2] { node [label={[label distance=-1mm]right:{\scriptsize $\sim G(s_0,\bar{a}_1)$}}] {$s_1$}
      child [norm] { node [action] {$\bar{a}_3$} }
      child [norm] { node [action] {$\bar{a}_4$} }
      child [emph3] { node [action] {$\bar{a}_5$}
        child [emph4, level distance=0.74in] { node [evaluate, label={[xshift=1.05cm, yshift=-0.35cm]above:{\scriptsize {\sc Rollout$(s_1,d)$}}}] {\scriptsize {\sc Evaluate$\left([s_0,s_1,\dots]\right)$}}
          edge from parent [evaluate-edge]
        }
      }
    }
  }
  child [norm, missing]
  child [norm] {node [action] {$\bar{a}_2$}
    child [norm] { node [label={[label distance=-1mm]right:{\scriptsize $\sim G(s_0,\bar{a}_2)$}}] {$s_2$} 
      child [norm] { node [action] {$\bar{a}_6$} }
      child [norm] { node [action] {$\bar{a}_7$} }
    }
  };
\end{tikzpicture}}
\caption{MCTS-PW with a single deterministic state and system evaluations at the end of the rollout.}
\label{fig:mcts_pw}
\end{figure}
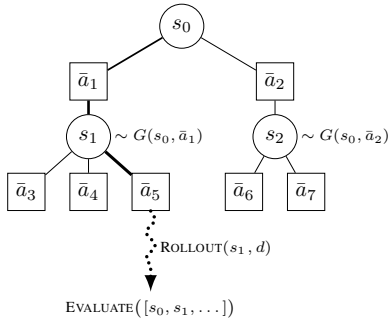

\subsection{Modified Monte Carlo Tree Search}
\label{sec:mcts}
Monte Carlo tree search (MCTS) is an anytime algorithm that uses rollouts of a random policy to estimate the value of each state-action node in the tree \cite{coulom2006efficient,dmubook}.
MCTS has found success in recent years in the reinforcement learning field, notably playing games such as Go \cite{silver2016mastering}.
There are four main stages in each simulation: \textit{selection}, \textit{expansion}, \textit{rollout} (or \textit{simulation}), and \textit{backpropagation}.
\Cref{fig:mcts} illustrates these four steps.
The algorithm is ``anytime'' because a policy can be constructed after any single iteration, but the state-action value estimates become increasingly accurate as more simulations are performed and the tree depth is expanded.
The tree $\mathcal{T}$ is iteratively expanded and the policy improves over time as the algorithm balances exploration with exploitation of the state and action spaces. 

\begin{algorithm}[!b]
  \caption{Top-level Monte Carlo tree search algorithm.} 
  \label{alg:mcts-pw}
  \begin{algorithmic}
  \Function{MonteCarloTreeSearch$(s, d)$}{}
  \Loop {}
    \State \textproc{Simulate}$(s,d)$
  \EndLoop
  \State \Return $\argmax\limits_{\bar{a} \in A(s)} Q(s, \bar{a})$
  \EndFunction
  \end{algorithmic}
\end{algorithm}
%
\begin{algorithm}[!b]
  \caption{Monte Carlo tree search simulation.} 
  \label{alg:mcts-pw-simulate}
  \begin{algorithmic}
  \Function{Simulate$(s, d)$}{}
  \If {$d=0$}
    \State \Return $0$
  \EndIf
  \If {$s \not\in \mathcal{T}$}
    \State $\mathcal{T} \leftarrow \mathcal{T} \cup \{s\}$
    \State $N(s) \leftarrow N_0(s)$
    \State \Return \textproc{Rollout}$(s,d)$
  \EndIf
  \State $N(s) \leftarrow N(s) + 1$
  \State $\bar{a} \leftarrow \textproc{SelectAction}(s)$ \algorithmiccomment{selection}
  \State $(s^\prime, r) \leftarrow \textproc{DeterministicStep}(s,\bar{a})$ \algorithmiccomment{expansion}
  \State $q \leftarrow r + \gamma \textproc{Simulate}(s^\prime, d-1)$ \algorithmiccomment{simulation/rollout}
  \State $N(s,\bar{a}) \leftarrow N(s,\bar{a})+1$
  \State $Q(s,\bar{a}) \leftarrow Q(s,\bar{a})+\frac{q-Q(s,\bar{a})}{N(s,\bar{a})}$ \algorithmiccomment{backpropagation}
  \State \Return $q$
  \EndFunction
  \end{algorithmic}
\end{algorithm}
%
\begin{algorithm}[!b]
  \caption{Modified rollout with end-of-depth evaluation.}
  \label{alg:mcts-pw-rollout}
  \begin{algorithmic}
  \Function{Rollout$(s,d)$}{}
  \If {$d = 0$}
    \State $(p, e, d) \leftarrow \textproc{Evaluate}(s)$
    \State $\tau \leftarrow \textproc{IsTerminal}(s)$
    \State $\bar{a}^* \leftarrow \textproc{UpdateBestAction}(s,Q)$
    \State \Return $R(p,e,d,\tau)$
  \ElsIf {$d = \floor{d_\text{max}/2}$}
    \State $\bar{a} \leftarrow \bar{a}^*$ \algorithmiccomment{feed best action}
  \Else
      \State $\bar{a} \leftarrow \textproc{SampleAction}(s,Q)$
  \EndIf
  \State $(s^\prime, r) \sim G(s,\bar{a})$
  \State \Return $r + \gamma$ \textproc{Rollout}$(s^\prime,d-1)$
  \EndFunction
  \end{algorithmic}
\end{algorithm}

A commonly used extension of MCTS for large or continuous spaces is progressive widening (PW) \cite{chaslot2007progressive,mcts_ucb,ll_wildfire,mcts_wildfire}.
We apply PW on the action space of seeds because there are an infinite number of seeds.
There is no need to apply PW to the state transitions, since a seed uniquely determines the next state. 
This notion can be seen in \Cref{fig:mcts_pw} where each action $\bar{a}_i$ leads to a single deterministic state $s_j$ as its child node.
The states are deterministically sampled from the generative model $G$ given the current state and action.
We define the state $s$ as a collection of all preceding actions deterministically leading to that point in the simulation.
Note that because the state $s$ is a sequence of seeds that uniquely determines the state of the simulator $\bar{S}$, we may overload the notation when calling $\smallcaps{Evaluate}$ and $\smallcaps{IsTerminal}$ for convenience.
In our formulation, the generative model samples state transitions and does not evaluate the underlying system (unlike standard MCTS).
This difference is in \Cref{alg:mcts-pw-rollout}, $\smallcaps{Rollout}$, where new state transitions are generated during the rollout and the system is only evaluated at the end.
By default, actions are uniformly selected from a random policy.
However, to encourage exploration of promising actions, the current best action $\bar{a}^*$ is used mid-rollout.
The best action is updated at the end of the rollout based on updated $Q$-values.

\begin{algorithm}[!t]
  \caption{Action selection with progressive widening.}
  \label{alg:mcts-pw-action-widen}
  \begin{algorithmic}
  \Function{SelectAction$(s)$}{}
  \If {$| A(s) | \le kN(s)^\alpha$}
    \State $\bar{a} \leftarrow \textproc{SampleAction}(s,Q)$
    \State $(N(s,\bar{a}), Q(s,\bar{a})) \leftarrow (N_0(s,\bar{a}), Q_0(s,\bar{a}))$
    \State $A(s) \leftarrow A(s) \cup \{\bar{a}\}$
  \EndIf
  \State \Return $\argmax\limits_{\bar{a} \in A(s)} Q(s,\bar{a}) + c\sqrt{\frac{\log N(s)}{N(s,\bar{a})}}$
  \EndFunction
  \end{algorithmic}
\end{algorithm}
\begin{algorithm}[!t]
  \caption{Single deterministic next state.} 
  \label{alg:mcts-pw-deterministic-state}
  \begin{algorithmic}
  \Function{DeterministicStep$(s,\bar{a})$}{}
  \If {$N(s,\bar{a},\cdot) = \emptyset$}
    \State $(s^\prime, r) \sim G(s,\bar{a})$
    \State $\textproc{SetCache}(s, \bar{a}, s^\prime, r)$
    \State $N(s,\bar{a},s^\prime) \leftarrow N_0(s,\bar{a},s^\prime)$
  \Else
    \State $(s^\prime, r) \leftarrow \textproc{GetCache}(s,\bar{a})$
    \State $N(s,\bar{a},s^\prime) \leftarrow N(s,\bar{a},s^\prime)+1$
  \EndIf
  \State \Return $(s^\prime,r)$
  \EndFunction
  \end{algorithmic}
\end{algorithm}

These modifications were made to MCTS-PW specifically for episodic reward problems.
The evaluation of our SUT is expensive so we limit the evaluations to the end of the rollout (rather than at node creation and throughout the rollout).
This way, we can reduce the number of external system executions but still provide the search with information during tree expansion, namely using back-propagated values of the transition probabilities and the miss distance from previously finished rollouts.
Choosing to evaluate at the end of the rollout provides the SUT with an expanded set of waypoints which it evaluates once the rollout has reached its maximum depth.
Generally for AST problem formulations, the discount factor $\gamma$ is set to $1$.
Algorithm \ref{alg:mcts-pw} is the entry point of MCTS-PW.

\section{Implementation}
\label{sec:implementation}
The AST implementation was written in the Julia programming language \cite{julia}.
Implementation of the simulation environment around the SUT was also written in Julia, but this section will focus on the algorithms required for AST and MCTS-PW.
Modifications to MCTS were implemented and merged into the existing MCTS.jl\footnote{https://github.com/JuliaPOMDP/MCTS.jl} Julia package.

\subsection{Interface}\label{sec:implementation_interface}
To apply AST to a general black-box system, a user has to provide the interface defined in \Cref{tab:interface}.
The simulation object $\bar{\mathcal{S}}$ is the user-defined data structure that holds parameters for their simulation.
All of the following functions take the simulation object $\bar{\mathcal{S}}$ as input and can modify the object in place.
The {\sc Initialize} function resets the simulation and the SUT to an initial state.
The {\sc Evaluate} function executes the SUT and returns the transition probability $p$, a boolean indicating an event occurred $e$, and the miss distance $d$.
Three subroutines determine these output values: {\sc Transition}, {\sc MissDistance}, and {\sc IsEvent} (where all three subroutines are used by the {\sc Evaluate} function, but may also be called individually).
Finally, the {\sc IsTerminal} function returns a boolean $\tau$ to indicate if the simulation is in a terminal state.

\begin{table}[!h]
  \centering
  \caption{\label{tab:interface} Adaptive Stress Testing Interface}
  \begin{threeparttable}
  \begin{tabular}{@{}p{3cm}l@{}} 
    \toprule
    \textbf{Function} & \textbf{$\bm{\text{Input}\mapsto\text{Output}}$} \\
    \midrule
    \textsc{Initialize} & $\bar{S} \mapsto \emptyset$ \\
    \textsc{Evaluate} & $\bar{S} \mapsto \langle p, e, d \rangle$ \\
    $\quad$\textsc{Transition} & $\bar{S} \mapsto p \in \mathbb{R}$ \\
    $\quad$\textsc{MissDistance} & $\bar{S} \mapsto d \in \mathbb{R}$ \\
    $\quad$\textsc{IsEvent} & $\bar{S} \mapsto e \in \mathbb{B}$ \\
    \textsc{IsTerminal} & $\bar{S} \mapsto \tau \in \mathbb{B}$ \\
    \bottomrule
  \end{tabular}
  \end{threeparttable}
\end{table}

As an example, the functions in the above interface can either be implemented directly in Julia or can call out to C++, Python, MATLAB\textsuperscript{\textregistered} or run an executable on the command line. Typically, implementing the {\sc MissDistance} and {\sc IsEvent} functions rely solely on the output of the SUT, thus keeping in accordance with the black-box formulation.


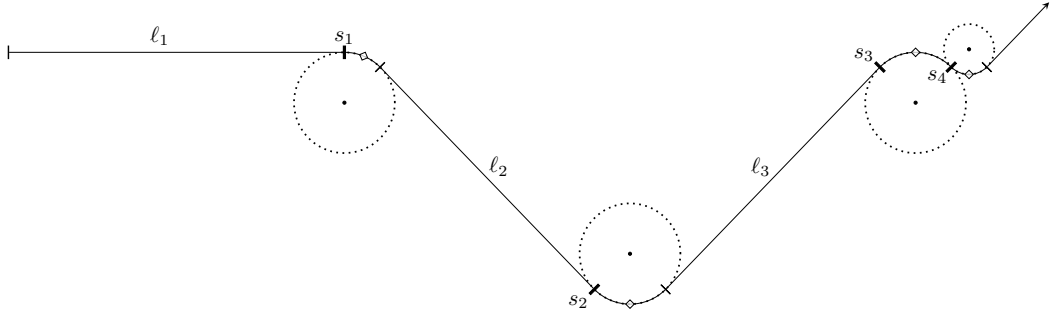
\begin{figure*}[!t]
\centering
\resizebox{0.78\textwidth}{!}{\begin{tikzpicture}
  [
    scale=16, every node/.style={scale=2},
    >=stealth,
    point/.style = {draw, circle,  fill = black, inner sep = 1pt},
    latlon/.style = {draw, rectangle, minimum width=0.0001mm, minimum height=2mm, fill = black, inner sep = 0.1pt},
    latlonstart/.style = {draw, rectangle, minimum width=0.0001mm, minimum height=2mm, fill = black, inner sep = 0.5pt},
    dot/.style   = {draw, circle,  fill = black, inner sep = .2pt},
    wpt/.style   = {diamond, fill=white, fill opacity=0.75},
  ]

  \tikzstyle{every node}=[font=\normalsize]

  \def\turnrad{0.05}
  \def\ang{45}

  \node (initial) at +(0,1/4+\turnrad) [latlon] {}; 
  \node (center) at +(1/3,1/4) [point, inner sep=0.5pt] {}; 
  \path (center) ++(90:\turnrad) node (start) [latlonstart, label = {[label distance=-1mm]above:$s_1$}] {};
  \path (center) ++(90-\ang:\turnrad) node (end) [] {}; 

  \draw (center) circle (\turnrad) [dotted,thick];

  \def\startang{90}
  \def\endang{90-\ang}
  \draw (start) arc (\startang:\endang:\turnrad);

  \path (center) ++(90-\ang/2:\turnrad) node (mid) [draw, rotate=-\ang/2, wpt, inner sep=1pt] {}; 

  \draw[-] (initial) -- node (ls) [label = {[label distance=-2mm]above left:$\ell_1$}] {} (start);

  \node [rotate=-\ang] (initial2) at (end) [latlon] {}; 
  \node (center2) at +(2/3+-\turnrad,2*\turnrad) [point, inner sep=0.5pt] {}; 
  \path (center2) ++(180+\ang:\turnrad) node [rotate=-\ang] (start2) [latlonstart, label = {[label distance=-1.5mm]below:$s_2$}] {};
  \path (center2) ++(-\ang:\turnrad) node (end2) [] {}; 

  \draw (center2) circle (\turnrad) [dotted,thick];

  \def\startang{180+\ang}
  \def\endang{180+3*\ang}
  \draw (start2) arc (\startang:\endang:\turnrad);

  \path (center2) ++(270:\turnrad) node (mid2) [draw, wpt, inner sep=1pt] {}; 

  \draw[-] (initial2) -- node (ls) [label = {[label distance=-3mm]above right:$\ell_2$}] {} (start2);

  \node [rotate=\ang] (initial3) at (end2) [latlon] {}; 
  \node (center3) at +(3/3-2*\turnrad,1/4) [point, inner sep=0.5pt] {}; 
  \path (center3) ++(90+\ang:\turnrad) node [rotate=\ang] (start3) [latlonstart, label = {[label distance=-1.5mm]above:$s_3$}] {};
  \path (center3) ++(90-\ang:\turnrad) node (end3) [] {}; 

  \draw (center3) circle (\turnrad) [dotted,thick];

  \def\startang{90+\ang}
  \def\endang{90-\ang}
  \draw (start3) arc (\startang:\endang:\turnrad);

  \path (center3) ++(90:\turnrad) node (mid) [draw, wpt, inner sep=1pt] {}; 

  \draw[-] (initial3) -- node (ls) [label = {[label distance=-3mm]above left:$\ell_3$}] {} (start3);

  \node (initial4) at (end3) [] {}; 
  \path (initial4) ++(\ang:\turnrad/2) node (center4) [point, inner sep=0.5pt] {}; 
  \path (center4) ++(180+\ang:\turnrad/2) node [rotate=-\ang] (start4) [latlonstart, label = {[label distance=-2mm]below:$s_4$}] {};
  \path (center4) ++(-\ang:\turnrad/2) node [rotate=\ang] (end4) [latlon] {}; 

  \draw (center4) circle (\turnrad/2) [dotted,thick];

  \def\startang{180+\ang}
  \def\endang{180+3*\ang}
  \draw (start4) arc (\startang:\endang:\turnrad/2);

  \path (center4) ++(270:\turnrad/2) node (mid) [draw, wpt, inner sep=1pt] {}; 

  \path (end4) ++(\ang+\ang*\turnrad/2:2*\turnrad) node (start5) [] {}; 

  \draw[->] (end4) -- node (ls) [] {} (start5);


\end{tikzpicture}}
\caption{Lateral packets output by the trajectory predictor. Lateral packets consist of latitude and longitude points that describe straight line segments $\ell_i$ and turning arc segments starting at $s_i$. Straight segments are optional which can result in multiple turn segments sequenced together, as seen at $s_3$ and $s_4$.}
\label{fig:lateral_packets}
\end{figure*}

\subsection{Stress Testing Julia Framework}
We have implemented the AST interface written in Julia as part of a new package called POMDPStressTesting.jl.\footnote{https://github.com/sisl/POMDPStressTesting.jl}
This package is inspired by work originally done in the AdaptiveStressTesting.jl\footnote{https://github.com/sisl/AdaptiveStressTesting.jl} package, but POMDPStressTesting.jl adheres to the \texttt{MDP} interface defined by the POMDPs.jl\footnote{https://github.com/JuliaPOMDP/POMDPs.jl} package  \cite{pomdps_jl}.
Thus, POMDPStressTesting.jl fits into the POMDPs.jl ecosystem, which is why it can use the MCTS.jl package as an off-the-shelf solver.
This design choice allows other Julia packages within the POMDPs.jl ecosystem to be used; this includes solvers, simulation tools, policies, and visualizations.

The intention of the POMDPStressTesting.jl package is to provide the user with a virtual black-box interface they must define, and provide the necessary AST algorithms to run the search. Future modifications will focus on inclusion of benchmark falsification problems from the literature  \cite{arch_comp}.

\section{Application}
\label{sec:application}

The primary goal of this work is applying the modified AST formulation to a trajectory predictor in a developmental commercial FMS.
The following sections will describe the input and output specification of the trajectory predictor and detail the investigated failure event and the associated miss distance.
We will also describe the simulation environment constructed to run the SUT.

\subsection{Trajectory Predictor}
The inputs of the trajectory predictor controlled by AST are the origin and destination airports, a set of intermediate waypoints, and the wind direction and magnitude at each waypoint.
The output of the trajectory predictor is a detailed flight path which provides predicted vertical data, predicted lateral data (i.e. lateral packets, illustrated in \Cref{fig:lateral_packets}), and other flight path information currently unused in this application.
The following failure event and miss distance is calculated by parsing the SUT output after each evaluation.

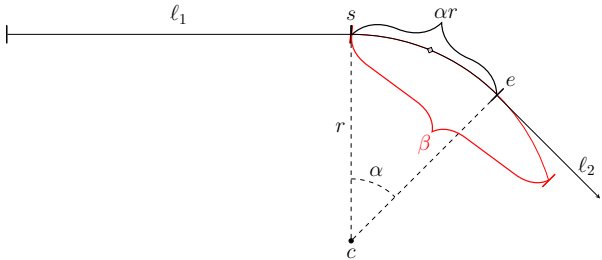
\begin{figure}[!hb]
\centering
\resizebox{0.9\columnwidth}{!}{\begin{tikzpicture}
  [
    scale=16, every node/.style={scale=2},
    >=stealth,
    point/.style = {draw, circle,  fill = black, inner sep = 1pt},
    dot/.style   = {draw, circle,  fill = black, inner sep = .2pt},
    latlon/.style = {draw, rectangle, minimum width=0.0001mm, minimum height=4mm, fill=white, fill opacity=0.75, fill = black, inner sep = 0.1pt},
    latlonstart/.style = {draw, rectangle, minimum width=0.0001mm, minimum height=4mm, fill = black, inner sep = 0.3pt},
    wpt/.style  = {diamond, fill=white, fill opacity=0.75},
  ]

  \tikzstyle{every node}=[font=\Large]

  \def\unit{0.5}
  \def\rad{0.3}
  \def\ang{45}
  \def\failang{7}
  \def\inner{0.3*\rad}
  \def\failinner{0.6*\rad}
  \def\failvalue{-28} 

  \node (origin) at (0,0) [point, label = {below:$c$}]{};

  \node (initial) at +(-\unit,\rad) [latlon] {};
  \node [rotate=-\failang] (failstart) at +(0, \rad) [latlonstart, red] {};
  \draw[red] (failstart) arc (90:\ang+\failvalue:\rad);

  \node (start) at +(0, \rad) [latlonstart, label = above:$s$] {};
  \node [rotate=-\ang] (end) at +(\ang:\rad) [latlon, label ={[label distance=-1mm]above:$e$}] {};
  \node (failend) at +(\ang+\failvalue:\rad) [rotate=-\ang, latlon,red] {};

  \draw (start) arc (90:\ang:\rad);

  \draw[-] (initial) -- node (ls) [label = {above:$\ell_1$}] {} (start);

  \draw[dashed]
    ($ (origin) ! 1 ! (start) $)
    -- ($(start) ! 1 ! (origin)$ )
    node (center) [right,label={[pos=0.45]left:$r$}] {};

  \draw[dashed]
    ($ (origin) ! 1 ! (end) $)
    -- ($(end) ! 1 ! (origin)$ )
    node [right] {};

  \draw[->, rotate around={\ang:(end)}] (end) -- node (nextstraight) [very near end, label = {above:$\ell_2$},] {} (end |-, 0);

  \draw [,thick,decorate,decoration={brace,amplitude=30pt}] (start) -- (end) node [black,midway,above=33pt,pos=0.65] {$\alpha r$};
  \draw [red,,thick,decorate,decoration={brace,amplitude=20pt}] (failend) -- (failstart) node [red,midway,below=27pt,pos=0.63] {$\beta$};

  \draw [dashed] (0, \inner) arc (90:\ang:\inner);
  \node (innerangle) at +(90-\ang+\ang/1.4:\inner) [label = {[label distance=-2mm]45:$\alpha$}] {};

  \node (mid) at +(90-\ang/2:\rad) [draw, wpt, inner sep=1pt, rotate=90-\ang/2] {};

\end{tikzpicture}}
\caption{Arc length $\beta$ and calculated arc length $\alpha r$, showing a failure in red.}
\label{fig:arc_length}
\end{figure}

\paragraph{Arc Length Failure}
Arc length is defined as the distance traveled across the arc from the starting point $s$ and ending point $e$, shown in \Cref{fig:arc_length}.
Failures can arise when the arc length $\beta$ does not agree with the arc length computed using the angular extent $\alpha$ and arc radius $r$.
Angular extent is computed as
\begin{equation}
  \alpha = \sign(r) \cdot |z_s - z_e| + 2\pi
\end{equation}
where $z_s$ is the azimuth from the center waypoint $c$ to the starting waypoint $s$, and $z_e$ is the azimuth from the center waypoint $c$ to the ending waypoint $e$.
The sign of $r$ determines the turn direction, where negative values represent left turns.
A failure occurs when the calculated difference $\abs{\beta - \alpha r}$ is above the threshold $h=10$ \si{ft}.

The arc length miss distance is how close the arc length difference comes to the threshold $h$.
We transform this difference by scaling the log-ratio of the threshold $h$ and the maximum miss distance from that trajectory.
This way, non-positive values indicate an event.
Namely, we define miss distance to be:
\begin{equation} \label{eq:miss}
  d = \rho\log\left(\frac{h}{\max\,\abs{\beta - \alpha r}}\right)
\end{equation}
We use a scale of $\rho=100$ to match the expected range of the log-likelihood in our problem so that the log-likelihood does not dominate the miss distance in the reward function.

\subsection{Simulation Environment}
A simulator was constructed to sample waypoint trajectories and evaluate the SUT.
Starting from an origin airport, waypoints were sampled from a multivariate Gaussian distribution of independent normals that encodes waypoint direction and distance (relative to the previous waypoint) and wind direction and magnitude, deterministically controlled by the seed $\bar{a}$:
\begin{equation}
  \vec{w} \overset{\bar{a}}{\sim} \Normal(\vec{\mu}, \vec{\Sigma})
\end{equation}
Mean and variance values for the waypoint direction and distance were set by a domain expert and the values for winds aloft were learned from observational weather data.
\begin{gather*}
\vec{\mu} = \left[180\,\si{\degree},\; 50\,\si{nmi},\; -88.5\si{\degree},\; 66.8\,\si{kts}\right]\\
\mat{\Sigma} = \begin{bmatrix}
45\si{\degree} & 0 & 0 & 0\\
0 & 30\,\si{nmi} & 0 & 0\\
0 & 0 & 39.5\si{\degree} & 0\\
0 & 0 & 0 & 24.4\,\si{kts}
\end{bmatrix}
\end{gather*}

The simulator implements the interface defined in \Cref{sec:implementation_interface} for the trajectory predictor.
The failure event and miss distance calculations described in \Cref{sec:application} were also implemented within the simulator.

\subsection{Navigational Database}
In addition to the simulator, we have access to a navigational database of aircraft routing procedures.
The routes are encoded as collections of waypoints describing departure, arrival, and en-route airways.
Exhaustively searching all combinations of the waypoints in the navigational database is the current approach to finding failures during development.
We employ the navigational database as a baseline by sampling the same allotted number of SUT evaluations to assess the miss distance distribution and search for failures.

\section{Experiments}
\label{sec:experiments}
Experiments were run to test the AST approach using MCTS-PW against direct Monte Carlo (MC) simulations as a na\"ive baseline and the cross-entropy method as an importance sampling baseline.
We also perform Monte Carlo sampling over the routes in the navigational database as another baseline.
Algorithm \ref{alg:mc} describes the direct Monte Carlo simulation approach for $n$ episodes, starting at an initial state $s_0$, with a rollout depth $d$. Note the rollout function does not use the action feeding procedure described in \Cref{sec:mcts}.

\algtext*{EndLoop}
\algtext*{EndIf}
\algtext*{EndFor}
\algtext*{EndFunction}
\begin{algorithm}[th!]
  \caption{Direct Monte Carlo simulation.}
  \label{alg:mc}
  \begin{algorithmic}
  \Function{DirectMonteCarlo$(s_0,n,d)$}{}
  \For {$1 \to n$}
    \State \textproc{Initialize}$(\bar{\mathcal{S}})$
    \State \textproc{Rollout}$(s_0, d)$ \algorithmiccomment{without action feeding}
  \EndFor
  \EndFunction
  \end{algorithmic}
\end{algorithm}

The cross-entropy method (CEM) is a probabilistic optimization technique that iteratively fits an initial distribution to elite samples \cite{rubinstein1999cross,rubinstein2004cross}.
The method uses importance sampling, which introduces a proposal distribution over rare events to sample from then re-weights the posterior likelihood by the \textit{likelihood ratio} of the true distribution ($\vec w$ in our case) over the proposal distribution.
The idea is to artificially make failure events less rare under the newly fit proposal distribution.
We set the proposal distribution to be the same as the true distribution $\vec w$, with the exception that the waypoint distance was reduced to $\mu=1\,\si{nmi}$ and $\sigma=3\,\si{nmi}$ to encourage smaller distances between the waypoints.

For the experiments, the San Francisco International Airport (KSFO) was used as an origin airport and the Los Angeles International Airport (KLAX) was used as a destination airport. 
Comparisons were run to assess the effectiveness of AST against CEM and direct MC in finding high-likely severe failure events. 
Metrics include the number of failure events found $N_E$, iteration of first failure $i_{FF}$, and statistics about the miss distance.
We also report the mean log-likelihood of failures found by each algorithm relative to the mean log-likelihood of failures found by the direct MC approach.
For a given algorithm, this is computed as:
\begin{equation}
 \operatorname{rel-log}(p) = \frac{\log(p_{\text{alg}})}{\log(p_\text{MC})}
\end{equation}
Values larger than one indicate a higher relative likelihood.
\begin{table}[!t]
  \centering
  \caption{\label{tab:mcts_params} Algorithm Hyperparameters}
  \begin{threeparttable}
  \begin{tabular}{@{}p{6.2cm}r@{}}
    \toprule
    \textbf{Hyperparameter} & \textbf{Value} \\
    \midrule
    episodes \tnote{*} & $5000$ \\
    maximum tree depth $d_\text{max}$ (i.e. number of waypoints) \tnote{*} & $12$ \\
    rollout depth $d$ \tnote{$\dagger$} & $12$ \\
    exploration constant $c$ & $10$ \\
    progressive widening $k$ & $10$ \\
    progressive widening $\alpha$ & $0.3$ \\
    \bottomrule
  \end{tabular}
  \begin{tablenotes}
      \item[*] {Used by all algorithms.}
      \item[$\dagger$] {Used by MCTS and direct Monte Carlo.}
  \end{tablenotes}
  \end{threeparttable}
\end{table}

All experiments were run with the MCTS hyperparameters listed in \Cref{tab:mcts_params}.
Sensitivity analysis of the various hyperparameter values has been omitted from this paper for brevity.
When controlling the parameters for progressive widening, to encourage widening let $k \to \infty$ and $\alpha \to 1$. To discourage widening, let $k \to 1$ and $\alpha \to 0$.

\subsection{Results and Analysis}

We first look at the performance of each approach over all episodes.
An initial seed is set across each experiment and we run each algorithm for 5000 episodes.
For each of these approaches, the number of episodes also corresponds to the number of SUT evaluation calls.

\begin{figure}[t!]
\centering
\resizebox{\columnwidth}{!}{\input{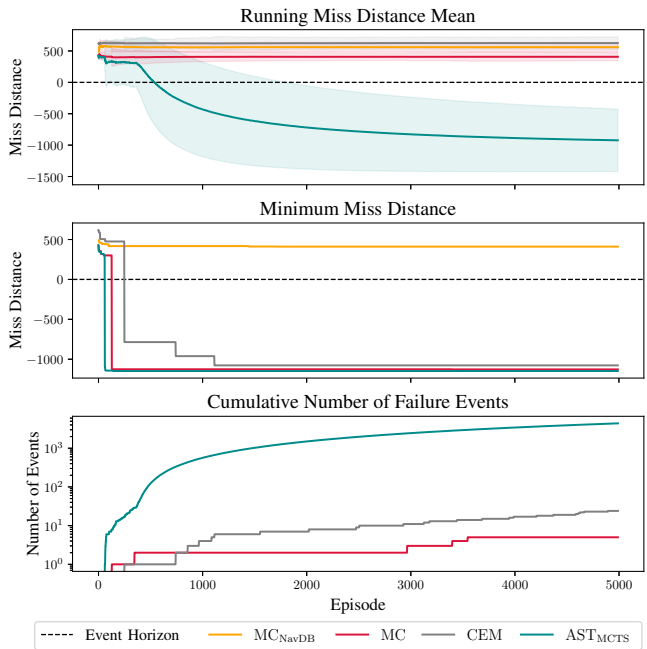}}
\caption{Running miss distance mean, minimum miss distance, and log-scaled cumulative number of failure events across episodes. One standard deviation is reported in the shaded regions.}
\label{fig:episodes}
\end{figure}

The first two plots in \Cref{fig:episodes} show the running mean and minimum miss distance over each episode.
Notice that the direct Monte Carlo approach applied to the navigational database baseline converges quickly to a minimum miss distance that remains above the rest, and a running mean that only outperforms the CEM approach.
Evident from \Cref{fig:episodes} is the initial behavior that MCTS and direct Monte Carlo share.
Recall that MCTS balances exploration and exploitation, and initially acts similar to direct Monte Carlo.
This similarity is based on the choice of exploration hyperparameters and the miss distance in the reward function.
This behavior suggests that the miss distance for this problem is a noisy measurement of the actual distance to a failure event.
At about episode 500, the MCTS approach starts to exploit found failures which can be seen as the descent of the running mean passing the origin (i.e. the event horizon).

One goal of the AST approach using MCTS is to exploit known failures to maximize their likelihood.
The bottom plot in \Cref{fig:episodes} shows the cumulative number of failure events which highlights this behavior.
Notice that each approach finds failures relatively early in the search, suggesting that failure events may be common given the choice of simulation environment.

\begin{figure}[t!]
\centering
\resizebox{\columnwidth}{!}{\input{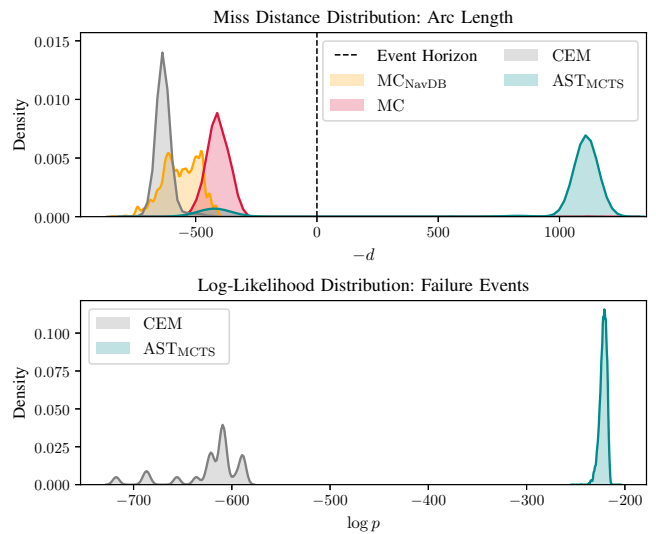}}
\caption{Distribution of negative miss distances for all episodes (where values to the right of the origin are events) and distribution of log-likelihoods filtered by failure events.}
\label{fig:distributions}
\end{figure}

We are also interested in the distribution of the miss distances collected from each approach.
Recall that the miss distance $d$ is a transformation of the arc length discrepancy relative to a threshold, detailed in \Cref{eq:miss}.
Thus, the value for $d$ is unitless and non-positive values indicate an event.
The top plot of \Cref{fig:distributions} shows the miss distance distributions, indicating the event horizon at the origin.
The miss distance distribution from the navigational database is used as a proxy for miss distance distributions we would expect in the real-world.
The CEM approach converges to a local minima and stays there, which is evident in the concentration of the CEM miss distance distribution.
MCTS and the direct MC approach share similar distributions to the left of the event horizon (indicating non-failure events), further suggesting that the miss distance is a noisy measurement of the true distance to a failure.
The spike to the right of the origin is the distribution of failure events found by our AST approach using MCTS.
The bottom plot in \Cref{fig:distributions} shows the distribution of log-likelihoods filtered for the failure events, suggesting that AST finds failures with higher likelihood than the CEM approach.

\begin{table}[!b]
  \centering
  \caption{\label{tab:results} Experimental Results}
  \resizebox{\columnwidth}{!}{%
  \begin{threeparttable}
  \begin{tabular}{@{}lrrrrr@{}}
    \toprule
    Algorithm\tnote{*} & $N_E$ & $i_{FF}$ & $\bar{X}_d$ & $\min(d)$ & $\operatorname{rel-log}(p)$\tnote{$\dagger$}\\
    \midrule
    $\text{MC}_\text{NavDB}$ & $0$          & ---        & $560.21 \pm75.09$       & $410.98$         & ---\\
    MC                       & $5$          & $128$        & $407.44 \pm64.85$       & $-1127.7$      & $1.0$\\
    CEM                      & $24$         & $249$        & $625.65 \pm97.80$        & $-1077.3$      & $4.5\times 10^{-161}$\\
    $\text{AST}_\text{MCTS}$ & $\bm{4394}$ & $\bm{61}$ & $\bm{-923.49 \pm497.4}$ & $\bm{-1147.9}$ & $\bm{13.1}$\\
    \bottomrule
  \end{tabular}
  \begin{tablenotes}
      \item[*] {Hyperparameters listed in \Cref{tab:mcts_params}.}
      \item[$\dagger$] {Mean log-likelihood relative to direct Monte Carlo.}
  \end{tablenotes}
  \end{threeparttable}}
\end{table}

The collected aggregate results are shown in \Cref{tab:results}.
AST finds failures with relative likelihood about an order of magnitude greater than that of direct Monte Carlo.
The CEM approach finds a small number of failures with very low relative likelihood.
This is because CEM is using importance sampling and after re-weighting the samples using the true distribution, we would expect to get these extremely small likelihood values.
AST has the lowest mean miss distance $\bar{X}_d$, noting the large standard deviation which is a result of large differences between miss distances from failure and non-failure events.
Each approach finds their first failure early in the experiment, with AST finding failures the earliest.
The effect of feeding the best action midway through the rollout accelerates finding these failures.
Once found, AST will exploit the failures to maximize their likelihood.
We see that AST finds failures in about $88\%$ of episodes (i.e. system executions), where as standard MC and CEM find failures in about $0.1\%$ and $0.48\%$ of episodes, respectively.

\subsection{Example Failure}

\begin{figure}[t!]
\noindent\makebox[\columnwidth][c]{%
\begin{minipage}{1.25\columnwidth}
\centering
\resizebox{0.8\columnwidth}{!}{\input{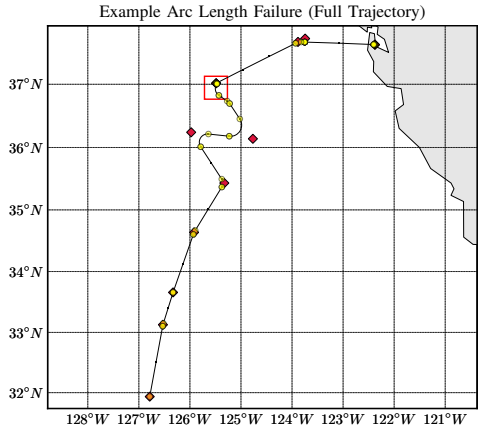}}
\end{minipage}}
\caption{Full trajectory of an example arc length failure originating from KSFO. Red diamonds indicate the input waypoints selected by MCTS and the yellow circles indicate the output lateral packets from the SUT. The red box shows where the failure occurs, shown in more detail in \Cref{fig:example_failure_zoomed}.}
\label{fig:example_failure}
\end{figure}

\begin{figure}[t!]
\noindent\makebox[\columnwidth][c]{%
\begin{minipage}{1.25\columnwidth}
\centering
\resizebox{0.8\columnwidth}{!}{\input{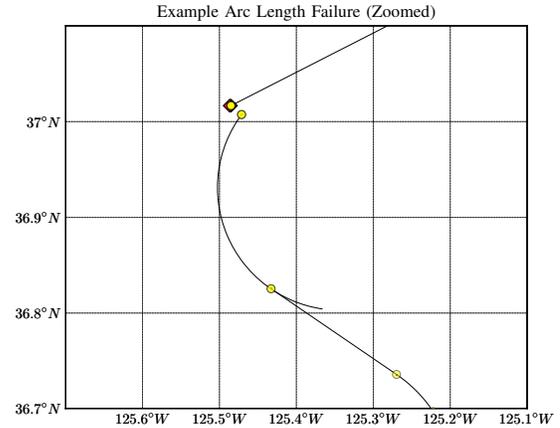}}
\end{minipage}}
\caption{Example arc length failure, zoomed in from \Cref{fig:example_failure}. Notice two almost identical red waypoint diamonds, which are separated by about 0.08 \si{nmi} or about 486 \si{ft} (zoom in further for more detail). The arc length failure is shown as the extending arc after the center yellow waypoint, which extends about 3 \si{nmi} past its intended end waypoint. This extension is the negative miss distance.}
\label{fig:example_failure_zoomed}
\end{figure}

Many of the failures are a result of two duplicate waypoints being generated in sequence.
Based on the defined environmental distributions this is possible, but unlikely.
However, certain failures found only by AST have waypoints close in range to each other---not necessarily identical---which can also result in arc length failures. 
\Cref{fig:example_failure} shows an example failure trajectory and \Cref{fig:example_failure_zoomed} zooms in on the specific arc length failure.
Refer to the figure captions for further descriptions.

Due to the nature of exploiting known failures, certain failure cases may only have minor differences between them.
To assess the impact of the trajectory predictor failures on broader flight operations, each failure case would have to be evaluated on the full FMS in simulation.
This would include modeling aircraft dynamics, guidance systems, and control feedback. 
Full assessment of the trajectory predictor failures would help inform the system developers in their decision to mitigate issues before deployment.
Further extensions of this work include searching for other failure events and stress testing other components of the FMS.

\section{Conclusion}
\label{sec:conclusion}
Adaptive stress testing was extended for sequential systems with episodic reward to find likely failures in FMS trajectory predictors.
To improve search performance, we used Monte Carlo tree search with progressive widening and modified the rollout with end-of-depth evaluations.
We feed the best action midway through the rollout to encourage exploration of promising actions, resulting in exploiting failures to maximize their likelihood. 
A simulation environment was constructed to evaluate the trajectory predictor, and a navigational database was sampled to compare to existing methods of finding failures during development.
Performance of AST using MCTS-PW was compared against direct Monte Carlo simulations and the cross-entropy method.
Results suggest that the AST approach finds more failures with both higher severity and higher relative likelihood.
The failure cases are provided to the system engineers to address unwanted behaviors before system deployment.
In addition to requirements-based tests, we show that AST can be used for confidence testing during development.

\section*{Acknowledgments}
The authors would like to thank GE's Global Research Center and GE Aviation for supporting this work through the Stanford Center for AI Safety.
We also thank the NASA AOSP System-Wide Safety Project for partially supporting this work.
We thank Anthony Corso and Mark Koren for their feedback and the Stanford Intelligent Systems Laboratory for their development of the POMDPs.jl ecosystem and the MCTS.jl package.
We would also like to thank Mark Darnell, Scott Edwards, and Andrew Foster for their technical support.

\printbibliography

\end{document}